\documentclass[10pt,twocolumn,letterpaper]{article}
\usepackage{colortbl}
\usepackage{amssymb}         
\usepackage[accsupp]{axessibility}

\usepackage[pagenumbers]{iccv} 

\usepackage[dvipsnames]{xcolor}

\usepackage{enumitem}
\usepackage{ulem}
\usepackage{graphicx}
\usepackage{amsmath}
\usepackage{amssymb}
\usepackage{arydshln} 
\usepackage{booktabs}
\usepackage{makecell}
\usepackage{color}
\usepackage{pifont}
\usepackage{float}
\usepackage{wrapfig}
\usepackage[caption=false]{subfig}
\usepackage{bm}
\usepackage{multirow} 
\usepackage{adjustbox}
\usepackage{bbding}
\makeatletter
  \newcommand\figcaption{\def\@captype{figure}\caption}
  \newcommand\tabcaption{\def\@captype{table}\caption}
\makeatother

\definecolor{annotation}{RGB}{0, 153, 0}
\definecolor{key_words}{RGB}{236, 0, 141}
\definecolor{RowColor}{rgb}{0.95, 0.95, 1}
\definecolor{cgray}{RGB}{220,220,220}
\definecolor{lightblue}{RGB}{163,199,235}
\definecolor{darkblue}{RGB}{0,76,153}
\definecolor{citegrey}{HTML}{75878a}
\definecolor{updatagreen}{RGB}{80,100,40}
\definecolor{updatagrey}{HTML}{686461}
\definecolor{citecolor}{HTML}{2980b9}
\definecolor{linkcolor}{HTML}{c0392b}
\definecolor{updategreen}{RGB}{80,100,40}
\definecolor{gtblue}{RGB}{50, 76,158}

\newcommand{\pub}[1]{{\color{citegrey}{\tiny{[{#1}]}}}}

\definecolor{iccvblue}{rgb}{0.21,0.49,0.74}

\usepackage[hidelinks,breaklinks=true,colorlinks,bookmarks=false,citecolor=citecolor,linkcolor=linkcolor]{hyperref}

\title{A Structure-aware and Motion-adaptive Framework for 3D Human Pose Estimation with Mamba}

\author{Ye Lu\textsuperscript{1*} \quad  Jie Wang \textsuperscript{2*} 
\quad Jianjun Gao\textsuperscript{1} \quad Rui Gong\textsuperscript{1}
\quad Chen Cai \textsuperscript{1} \quad Kim-Hui Yap\textsuperscript{1} \\[1.5ex]
    \textsuperscript{1~}Nanyang Technological University \qquad
    \textsuperscript{2~}Beijing Institute of Technology\\[1.1ex]
	{\tt\footnotesize \{lu0001ye@e.,gaoj0018@e.,gong0084@e.,e190210@e.,ekhyap@\}ntu.edu.sg} \quad
    {\tt\footnotesize \{jwang991020\}@gmail.com}\\}

\makeatletter
\def\sf@counterlist{}
\makeatother
    
\begin{document}
\maketitle
\begin{abstract}
\def\thefootnote{*}\footnotetext{Equal contribution}\def\thefootnote{\arabic{footnote}}
Recent Mamba-based methods for the pose-lifting task tend to model joint dependencies by 2D-to-1D mapping with diverse scanning strategies.
Though effective, they struggle to model intricate joint connections and uniformly process all joint motion trajectories while neglecting the intrinsic differences across motion characteristics.
In this work, we propose a structure-aware and motion-adaptive framework to capture spatial joint topology along with diverse motion dynamics independently, named as SAMA. Specifically, SAMA consists of a Structure-aware State Integrator (SSI) and a Motion-adaptive State Modulator (MSM). 
The Structure-aware State Integrator is tasked with leveraging dynamic joint relationships to fuse information at both the joint feature and state levels in the state space, based on pose topology rather than sequential state transitions.
The Motion-adaptive State Modulator is responsible for joint-specific motion characteristics recognition, thus applying tailored adjustments to diverse motion patterns across different joints.
Through the above key modules, our algorithm enables structure-aware and motion-adaptive pose lifting.
Extensive experiments across multiple benchmarks demonstrate that our algorithm achieves advanced results with fewer computational costs.
\end{abstract}    
\section{Introduction}

\begin{figure}[h]
\centering
\includegraphics[width=0.99\linewidth] {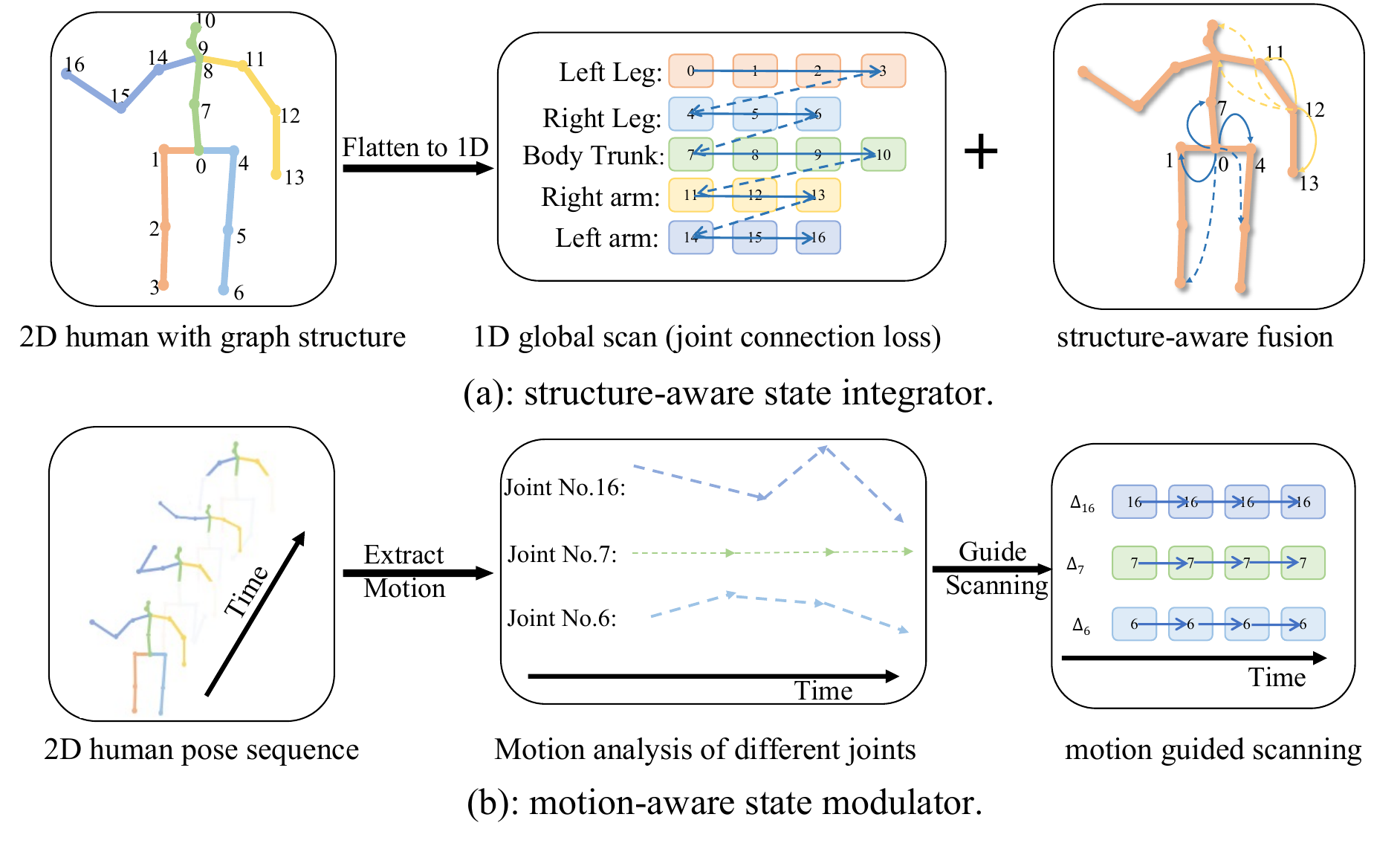}
\setlength{\abovecaptionskip}{-10pt}  
\setlength{\belowcaptionskip}{0pt}
 \caption{(a) Illustration of structure-aware state integrator. On top of the linear scanning, we aggregate joints based on their connections, supplementing the necessary learnable topology information. (b) Representation of motion-aware modulator.
We identify the distinct motion characteristics of different joints and adaptively generate timescales $\Delta$ to guide the model in capturing the unique motion features of these joints.}
  \label{fig:motivation}
\vspace{-12pt}
\end{figure}
\label{sec:intro}

Monocular 3D Human Pose estimation is a fundamental computer vision task, aiming to estimate 3D human poses in 3D space from single-view 2D images or videos. This technique serves as the foundation for a diverse range of applications, including action recognition \cite{DBLP:conf/cvpr/ZhangYWCWZ22,DBLP:journals/caaitrit/ZhangYTQQZL22} and human-computer interaction \cite{DBLP:conf/iccv/Chen0GZCY19,DBLP:journals/tip/ChenTKCBZY21,DBLP:journals/tip/YeLDSSH22}.
Approaches to this task generally fall into two categories: directly estimating 3D poses from images or videos \cite{DBLP:conf/iccv/LiZC15,DBLP:conf/eccv/MoonL20,DBLP:conf/aaai/ChenWL21,DBLP:conf/iccv/WehrbeinRRW21}, detecting 2D poses with off-the-shelf detectors and lifting them into 3D. Due to its more dependable performance, the 2D-to-3D pose lifting has become the mainstream based on robust 2D pose estimators.
However, monocular 2D pose often suffers from depth ambiguity, where one single 2D pose can correspond to multiple 3D poses, making it difficult to accurately recover 3D poses from a single frame of 2D keypoints.
Current methods address this issue by leveraging temporal information from videos to capture joint dependencies across space and time, achieving significant progress.

Recently, Mamba-based methods \cite{huang2024posemamba,DBLP:journals/corr/abs-2408-02922} have been introduced to the pose-lifting task using state space models \cite{gu2021combining,DBLP:journals/corr/abs-2312-00752,DBLP:conf/icml/DaoG24}, leveraging their linear complexity and effectively capturing detailed spatio-temporal joint dependencies.
Despite employing different scanning methods \cite{huang2024posemamba,DBLP:journals/corr/abs-2408-02922}, these approaches have limitations in effectively capturing complex joint interactions. 
Their uniform treatment of joint trajectories tends to overlook the inherent variations in motion patterns across different joints, as shown in \cref{fig:motivation}.
In the spatial domain, human joints are naturally connected by a specific graph structure, where each joint maintains connections with a varying number of neighbor joints.
Simply flattening this graph-structured pose into 1D data disrupts its inherent topology, resulting in the loss of crucial structural information and ultimately degrading pose estimation performance.
In the temporal domain, joint motions vary significantly, with arms and legs exhibiting high flexibility and large ranges, while the trunk remains more constrained.
Previous methods process all joint motion trajectories uniformly, ignoring their intrinsic motion differences, resulting in insufficient learning and suboptimal motion representation.
Thus, preserving pose topology and adaptively capturing joint-specific motion dynamics remains a challenge in these Mamba-based methods.

To address these limitations, we propose a structure-aware and motion-adaptive framework named as SAMA, as shown in \cref{fig:motivation}. 
It contains a structure-aware state integrator that efficiently fuses dynamic joint relations into the state space. Additionally, it includes a motion-adaptive state modulator to model joint-specific motion dynamics.
To incorporate structure-aware joint relationships, the proposed SSI fuses dynamic pose topology within both joint features and states in the state space. Specifically, we introduce a learnable adjacency matrix that encodes both the inherent joint connectivity and the learned global dependencies. This matrix guides the construction of a structure-aware embedding to enhance pose representation and facilitates state fusion in the state space. By integrating structural features, SSI mitigates the limitation of conventional state-space models that rely solely on sequential reasoning.
To capture joint-specific motion dynamics, our MSM adaptively regulates the timescale in the SSM, enabling the model to effectively adjust to varying motion patterns across joints.  
Specifically, it aggregates neighboring frame joint features to learn a joint-specific timescale, which adapts the model's reliance on the previous joint state and current joint input based on the unique motion characteristics of each joint. 
This adaptive dependency allows MSM to dynamically model diverse joint motion patterns.
By integrating SSI and MSM, our model captures the intrinsic connectivity between joints and adaptively learns the motion trajectory characteristics of different joints, achieving significant performance gains with minimal computational costs.

\begin{figure}[!t]
\setlength{\abovecaptionskip}{0pt}  
\setlength{\belowcaptionskip}{0pt}
\includegraphics[width=0.99\linewidth]
{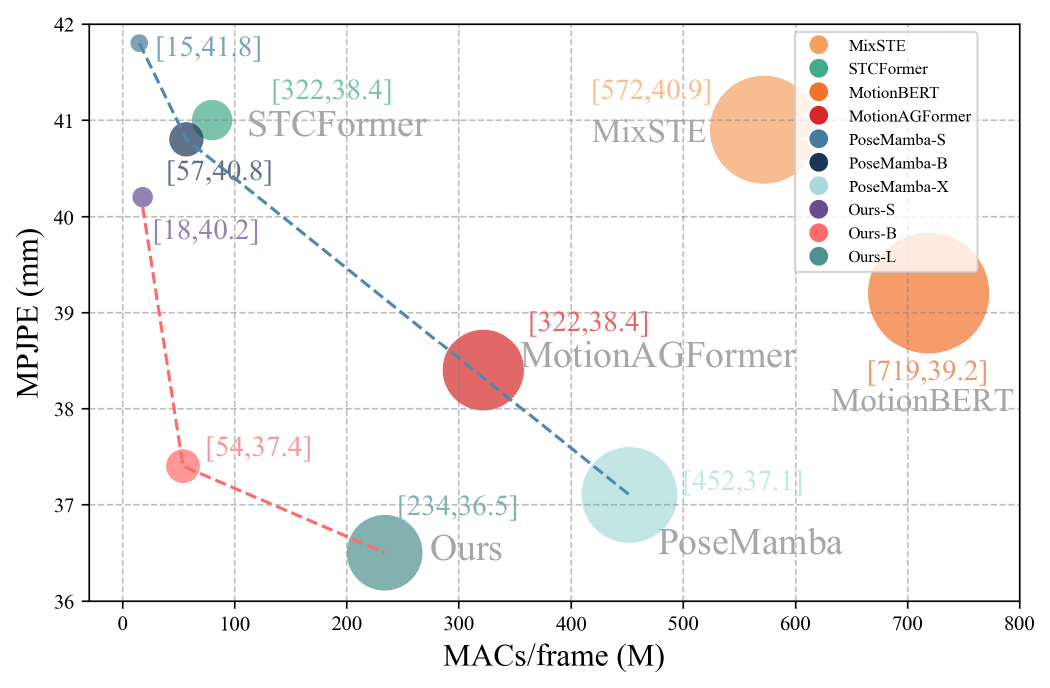}
  \caption{Comparisons of various 3D Human Pose Estimation methods on Human3.6M ($\downarrow$). MACs/frame represents multiply-accumulate operations per output frame. Radius denotes the parameters. Our method achieves superior results with fewer parameters and computation costs.}
\vspace{-0.5cm}
\label{fig:comparison fig in intro}

\end{figure}

We have extensively validated the effectiveness of our proposed method on multiple datasets, including Human3.6M and more challenging in-the-wild MPI-INF-3DHP. 
Our method surpasses the previous state-of-the-art (SOTA) methods with fewer parameters and MACs, as shown in ~\cref{fig:comparison fig in intro}.
Our experiment results also demonstrate that the proposed modules, SSI and MSM, improve the performance of diverse models, showing their generalization.
Our contributions can be summarized as follows:
\begin{itemize}
    \item We present a new framework, SAMA, which incorporates dynamic joint relations into the state space and captures joint-specific motion dynamics.
    \item We propose a method that adaptively captures spatiotemporal dependencies and dynamically adjusts the timescale for modeling joint-specific motion dynamics, based on local motion patterns through SSI and MSM.
    \item We demonstrate the effectiveness of SAMA  through extensive experiments across diverse datasets.
\end{itemize}

\section{Related Work}
\label{sec:formatting}

\subsection{2D-to-3D Pose Lifting}
Monocular 3D human pose estimation can be divided into two categories: direct 3D human pose estimation and 2D-to-3D pose lifting. Direct regression methods predict 3D human poses from 2D images or videos. 
End-to-end approaches \cite{DBLP:conf/cvpr/TekinRLF16,DBLP:conf/cvpr/PavlakosZDD17a,DBLP:conf/eccv/SunXWLW18} directly regress 3D poses from images or other raw data but require high computational costs and yield suboptimal results due to operating directly in the image space.
In contrast, 2D-to-3D pose lifting methods, which first detect 2D poses and then reconstruct 3D poses from these estimations, have demonstrated superior performance over direct regression approaches.
The existing pose lifting methods are classified into two types: Transformer-based methods and GCN-based methods. 
Transformers \cite{li2023pose,zheng2025hipart,lu2024hdplifter} are extensively used in pose-lifting tasks for capturing spatial and temporal joint correlations, leveraging their strong global modeling ability. 
PoseFormer \cite{DBLP:conf/iccv/ZhengZMY0D21} is the first to employ spatial and temporal Transformers separately to capture intra-frame joint dependencies and pose correlations across different frames. 
MixSTE \cite{DBLP:conf/cvpr/Zhang0YCY22} is a sequence-to-sequence model that alternates between spatial and temporal blocks to capture joint dependencies, and it proposes separately modeling the temporal correlations of different joints. 
GCN-based methods leverage the connection of human joints through bones, establishing essential spatial constraints and temporal coherence.
SemGCN \cite{DBLP:conf/cvpr/00030TKM19} proposes learning the relationships between directly connected joints and joints that are not physically connected, taking into account dynamic poses across various datasets and real-world applications. 
In GraFormer \cite{DBLP:conf/cvpr/ZhaoWT22}, the ChebGConv block was introduced to enable information exchange among nodes that lack direct connections, thereby capturing subtle relationships that may not be readily apparent. 
Overall, Transformer-based methods \cite{liu2023bitstream,cai2024empowering} face challenges to model pose structure and suffer from quadratic complexity, while GCN-based methods lack global modeling capability. 
In this manuscript, we introduce a novel Mamba-based approach that not only captures the dynamic structure of poses but also incorporates global modeling capabilities.

\subsection{Mamba-based Models in Human-Centric Tasks}
Mamba \cite{DBLP:journals/corr/abs-2312-00752} achieves Transformer-like capabilities with linear complexity by incorporating a data-dependent selective mechanism and a hardware-aware algorithm to facilitate highly efficient training and inference processes.
Based on that, Mamba2 \cite{DBLP:conf/icml/DaoG24} reveals the connections between SSMs and attention with specific
structured matrix and explore larger and more expressive state spaces through introducing State Space Duality. In human centric tasks, SSMs have been widely utilized with their strong global modeling ability and linear complexity. 
Motion Mamba \cite{DBLP:conf/eccv/ZhangLRHZT24} enhances temporal and spatial modeling, while Hamba \cite{DBLP:conf/nips/DongCG0T24} integrates graph learning with SSMs for structured joint relations.
For 2D-to-3D pose lifting task, previous works have leveraged state-space models to model spatiotemporal joint dependencies.
PoseMamba \cite{DBLP:journals/corr/abs-2408-03540} proposes a global-local spatio-temporal modeling approach within Mamba framework to address the 2D-to-3D pose lifting task. Posemagic \cite{DBLP:journals/corr/abs-2408-02922} propose a attention-free hybrid spatiotemporal architecture adaptively combining Mamba with GCN. However, these methods merely apply Mamba to the 2D-to-3D pose lifting task without accounting for the unique motion characteristics of human pose sequences and the inherent connections between joints in state space. In this manuscript, we introduce the structure-aware state integrator and the motion-adaptive state modulator to enhance Mamba's ability to capture the unique motion patterns of human pose sequences and the intrinsic connections between joints in state space.

\section{Method}
\subsection{Preliminaries}
\begin{figure*}[tp]
  \centering
  \includegraphics[width=0.99\linewidth] {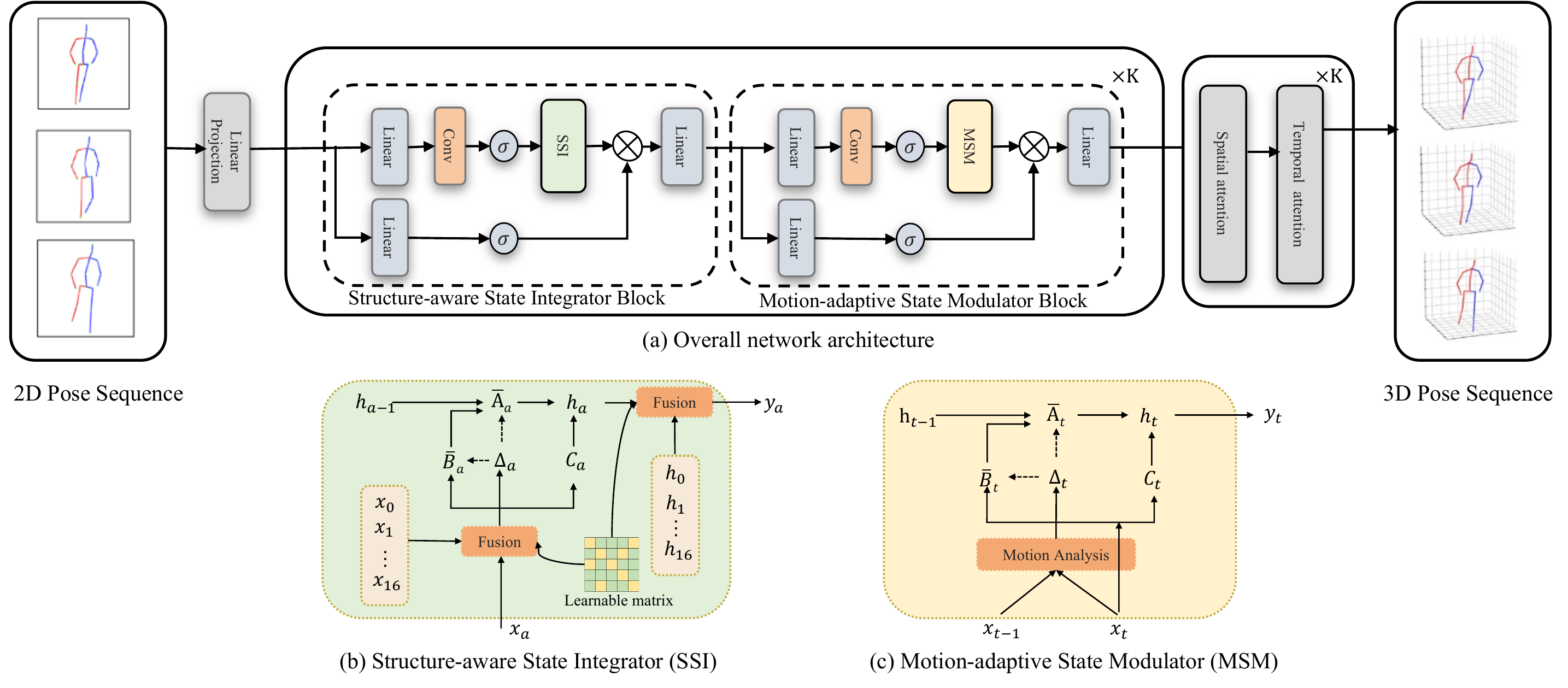}
 \vspace{-8pt}
 \caption{The overview of our proposed SAMA.
 (a): Our Network Structure. The core part is the alternative stack of structure-aware state integrator block and motion-adaptive state modulator block.
 (b): Our structure-aware state integrator with structure-aware fusion in state space.
 (c): Our motion-adaptive state modulator with adaptively joint motion modeling.}
  \label{fig:structure}
\vspace{-12pt}
\end{figure*}

\textbf{Mamba in pose lifting.} SSMs are widely applied in sequential data analysis and the modeling of continuous linear time-invariant (LTI) systems. This dynamic system can be described by the
linear state transition and observation equations:
$h'(t)=Ah(t)+Bx(t),y(t)=Ch(t)+Dx(t)$, where 
$A\in \mathbb{C}^{N\times N},B,C\in \mathbb{C}^{N},D\in \mathbb{C}^{1}$ are trainable parameters, $x(t)$ denotes the input sequence, $y(t)$ means the output sequence, and $h(t)$ represents state variable.

In the pose lifting task, the input is a sequence of 2D discrete poses $C_{n,t}\in \mathbb{R}^{N\times T \times 2}$ and the output is a sequence of 3D discrete poses $O_{n,t}\in \mathbb{R}^{N\times T \times 3}$, where $N$ denotes the number of joints in a single frame, and $T$ signifies the total number of frames.
To adapt SSMs to this discrete sequence input in the deep learning framework, PoseMamba utilized the Zero-Order Hold (ZOH) discretization, following the setting of Mamba. 
It discretizes the continuous-time system by assuming the input remains constant within each time interval and introducing a timescale $\Delta$ which represents the interval between adjacent timesteps. The ZOH method is applied to compute the discrete system parameters as follows: $\overline{\mathbf{A}} = e^{\Delta \mathbf{A}},\overline{\mathbf{B}} = (\Delta \mathbf{A})^{-1} (e^{\Delta \mathbf{A}} - I) \Delta \mathbf{B}.$

In addition, PoseMamba follows context-aware and adaptive SSMs in Mamba through modifying the parameter $\Delta,\overline{\mathbf{B}},\overline{\mathbf{C}}$ as functions of the input sequence $x_t$, resulting a data-independent parameters ${\boldsymbol{\Delta}}_t=s_{\Delta}(x_t)$, $\overline{\mathbf{B}}_t = s_B(x_t)$, and $\mathbf{C}_t = s_C(x_t)$. 
Following previous methods \cite{DBLP:conf/cvpr/Zhang0YCY22,DBLP:conf/cvpr/Li00WG22,DBLP:conf/iccv/ZhuMLLW023}, PoseMamba models spatial and temporal joint dependencies separately. 
In the spatial modeling, PoseMamba processes joints feature in one frame $X_n \in \mathbb{R}^{N\times d}$, where $d$ denotes the dimension of features. The discrete spatial state transition equation and observation equation are formulated as:
\begin{equation}
h_{n} = \overline{\mathbf{A}}_{n} h_{n-1} + \overline{\mathbf{B}}_{n} x_{n}, \quad y_{n} = \mathbf{C}_{n} h_{n}.
\label{ssm}
\end{equation}

In the temporal modeling, PoseMamba processes joints feature in a joint motion trajectory  $X_t \in \mathbb{R}^{T\times d}$. The discrete version of the temporal state transition equation and observation equation is similar to \cref{ssm}.

\noindent \textbf{Mamba2 and State Space Duality.}
Based on Mamba, Mamba2 draws connection between SSMs and Transformers by introducing Structured State Space Duality (SSD). Different from Mamba1, Mamba2 restrict $\overline{A}=\alpha_t * I$, where $I$ denotes Identity Matrix, leading to the formulation of causal linear attention.
Due to the aforementioned connection between SSMs and Transformers,
the SSD mixer family of Mamba-2 has been shown to be equivalent to sequentially-semi-separable matrices. The SSD can be expressed as:
\begin{equation}
\begin{aligned}
    h_{t} &=\overline{A} h_{t-1} + \overline{B} x_t, \quad y_t = {C} h_t,
\end{aligned}
\label{mamba2}
\end{equation}
The quadratic form of \cref{mamba2} can be reformulated as:
\begin{equation}
    y_t = P \cdot (C^T B)x, \quad 
\label{eq:M_definition}
\end{equation}
where $P_{ij}$ is defined as follows: $P_{ij} = \overline{A}_{i+1} \times \cdots \times \overline{A}_j$ if $i > j$, $P_{ij} = 1$ if $i = j$, and $P_{ij} = 0$ if $i < j$. Hence, Mamba2 network is regarded as a causal linear attention with a learnable causal mask. In this work, we employ the SSD in Mamba2 as the baseline to construct SAMA due to its training stability and ease of implementation.

\subsection{Overall Architecture}

As illustrated in \cref{fig:structure} (a), our network processes a 2D pose sequence $C_{n,t}\in \mathbb{R}^{N\times T \times 2}$ and outputs a 3D pose sequence $O_{n,t}\in \mathbb{R}^{N\times T \times 3}$.
Firstly, a linear projection layer is used to project the input into high dimension feature  $X\in \mathbb{R}^{N\times T \times d}$. 
In contrast to previous methods, the spatial and temporal position embeddings are not added to the high-dimensional features, because models such as SSM are already capable of capturing token positional order, making the additional positional information redundant.
Next, several layers of structure-aware state integrator and motion-adaptive state modulator capture dynamic spatial and temporal joint correlations in an alternating manner. SSI is designed to enable the fusion of joint features and hidden states among joints. Meanwhile, MSM considers the differences in motion characteristics among joints by learning the timescale from the joint motion information to dynamically learn each joint's unique motion properties.

\subsection{Structure-aware State Integrator}
Structure-aware state integrator is designed to effectively capture the spatial dependencies between adjacent joints within the latent state space, as shown in \cref{fig:structure} (b). To achieve this goal, unlike previous methods that repeatedly scan using different approaches, we introduce a structure-aware state transition into original Mamba formulas. We first construct a learnable matrix to dynamically model the relationships between joints. Then, we use the designed matrix to aggregate joint features and state information.

\noindent\textbf{Construction of the learnable adjacency matrix.}
To efficiently model joint connections in the state space, a learnable adjacency matrix $M$ is defined as follows:
\begin{equation}
\vspace{-3pt}
M=\text{softmax}( D^{-\frac{1}{2}}\left(\ M_o+I \right)D^{-\frac{1}{2}}),
\label{adj}
\vspace{-3pt}
\end{equation}
where $D$ denotes the degree of each joint and I represents the identity matrix. $M_o\in \mathbb{R}^{N \times N}$ means the adjacency matrix and $M\in \mathbb{R}^{N \times N}$ represent a learnable adjacency matrix with global perception and enhanced attention to the connected joint. 
In \cref{adj}, we normalize the adjacency matrix based on joint degrees, as different joints have varying connections. Given the diversity of human actions, we set the normalized adjacency matrix as a learnable parameter to adapt to this variability. Additionally, \cref{adj} provides an initialization for $M$.

\noindent\textbf{Structure-aware joint feature and state fusion.}
By using the learnable adjacency matrix, we can achieve the fusion of joint features and states. Since the aggregation is implemented using an $M\in \mathbb{R}^{N \times N}$
matrix, we can save more computational cost compared to the previous method of repeated scanning.
The process of the structure-aware joint feature and state fusion can be described by four equations: the joint feature fusion equation, the state transition equation,
the structure-aware state fusion equation and the observation equation.
In the joint feature fusion equation, we first add structure-aware information to the input through the learnable matrix in \cref{adj}:
\vspace{-5pt}
\begin{equation}
x_a' = x_a + \sum_{k=0}^{N-1} M_{ak} x_{k}
\vspace{-5pt}
\label{structure-feature}
\end{equation}
\noindent where $x_a$ is the feature of $a$-th joint, $x_a'$ is the feature of $a$-th joint after structure-aware joint fusion.
Then, we compute the state $h_a$ based on the state transition equation: $h_{a} = \bar{A}_a h_{a-1} + \bar{B}_a x_a'$. In addition, we also update the hidden state of joints by incorporating other joint hidden states through the adjacent matrix with the structure-aware state fusion equation :
\begin{equation}
H_a = h_a+\sum_{k=0}^{N-1} M_{ak} h_k
\label{state-fusion}
\end{equation}
\noindent where, $h_a$ is the original hidden state, $H_a$ is the structure-aware hidden state.
Finally, we employ the observation equation: $y_{a}=C_a H_a$, where $y_{a}$ is the output feature of $(a)$-th joint. 
Compared with \cref{ssm}, we can observe that the joint feature and hidden state are directly influenced by other joints, especially the connected joints. However, in previous methods \cite{huang2024posemamba,DBLP:journals/corr/abs-2408-02922}, the current joint could only be influenced by joints with a smaller index in the scan.

\subsection{Motion-adaptive State Modulator}
The previous Mamba-based method, when modeling the temporal motion of joints, ignored the differences in motion characteristics among different joints and simply fed the raw joint trajectories into the SSM.
MSM is designed to adaptively learn the motion characteristics of different joints, capturing their unique dynamics and improving motion representation, as shown in \cref{fig:structure} (c).
We first propose capturing the motion characteristics of different joints and using these characteristics to dynamically learn the timescale, which controls the model's reliance on the current input and previous state. Then, we introduce two simple methods to model the timescale based on motions.

\noindent \textbf{Motion-aware timescale.}
The timescale $\Delta$, which controls the balance between how much to focus or ignore the current input, is an important parameter in Mamba and Mamba2. Typically, the timescale is designed as a learnable parameter determined by each token in other tasks. However, the joint motion trajectories exhibit different characteristics across different joints. Specifically, joints in the legs and arms exhibit high motion intensity, so a larger timescale should be used at certain moments to focus on the current input. On the other hand, joints in the body trunk have lower motion intensity, so a smaller timescale should be used to maintain continuity and preserve the state.
Different from the previous method, which ignores the motion characteristics of different joints, we use the features of adjacent joints as input to learn the timescale:
\begin{equation}
\begin{aligned}
\Delta_t &= S_{\Delta}(x_t, x_{t-1})
\end{aligned}
\end{equation}
where $S_{\Delta}$ denotes a learnable function, with $x_t$ and $x_{t-1}$ representing the features of the same joint at adjacent time steps.
This design enables the timescale to adapt dynamically to varying joint motion characteristics, ensuring a more flexible and responsive modeling of joint dynamics.

\noindent  \textbf{Practical implementation.}
We employ two different functions to model the timescale $\Delta$: point-wise convolution, and linear transformation. 
For the point-wise convolutions, we use a kernel size of 2 in the temporal dimension and apply zero padding at the start to capture local motion patterns. For the linear transformation, we concatenate adjacent joint features along the feature dimension, with zero padding applied at the start to preserve all the features.

\subsection{Network Architecture.}
The overall architecture is illustrated in \cref{fig:structure} (a). We alternately stack structure-aware state integrator and motion-adaptive state modulator for $K$ layers.
Following Jamba \cite{lieber2024jamba}, we integrate $K$ layers of spatial and temporal attention to further enhance joint correlation modeling.

\subsection{Overall Learning Objectives}
Following the previous method \cite{DBLP:conf/iccv/ZhuMLLW023}, we train the model with a end-to-end manner.
The final loss is defined as:
\begin{equation}
\mathcal{L} = \mathcal{L}_w + \lambda_m \mathcal{L}_m + \lambda_n \mathcal{L}_n,
\label{eq:loss_function}
\end{equation}
where \({L}_w\) is weighted MPJPE,   
$L_m$ denotes MPJVE, and $L_n$  represents Normalized MPJPE. We set $\lambda_m$ to 20 and $\lambda_n$ to the default value of 0.5, respectively.

\section{Experiments}
\label{Experiments}
We first introduce the experimental setup in \S\ref{para:Experiments - Experimental Setup}. Then we assess the performance of our method 
across various datasets, including indoor Human3.6M in~\S\ref{para:Experiments - Indoor monocular 3D human pose estimation}, and more challenging in-the-wild datatset MPI-INF-3DHP in~\S\ref{para:Experiments - In-the-wild pose estimation}. Lastly, we provide ablative analyses in ~\S\ref{para:Experiments - Ablation Study}.

\begin{table*}[!ht]
    \centering
    \setlength{\tabcolsep}{10.0pt}
    \renewcommand\arraystretch{1.0}
    \setlength{\abovecaptionskip}{5pt}  
    \setlength{\belowcaptionskip}{5pt}
    \caption{Quantitative comparisons on Human3.6M.
    $T$: Number of input frames. CE: Estimating center frame only. MACs/frame: multiply-accumulate operations per output frame. P1: MPJPE (mm). P2: P-MPJPE  (mm). ${\mathrm{P1}^\dag}$: P1 on 2D ground truth. (*) denotes using HRNet for 2D pose estimation. The best and second-best scores are in bold and underlined, respectively. }
    \label{tab:human3.6m-comparison}
    \small
    \begin{tabular}{lccccccc}
      \toprule
      Method & $T$ & CE & Param(M) & MACs(G) & MACs/frame(M) & P1$\downarrow$	
/P2$\downarrow$	 & ${\mathrm{P1}^\dag}$$\downarrow$	 
 \\ 
      \midrule
      *MHFormer \pub{CVPR2022} \cite{DBLP:conf/cvpr/Li00WG22}& 351 & \checkmark & 30.9 & 7.0 & 20 & 43.0/34.4 & 30.5 \\
      Stridedformer \pub{TMM2022} \cite{DBLP:journals/tmm/LiLDLWY23} & 351 & \checkmark & 4.0 & 0.8 &2 &43.7/35.2 & 28.5\\
      Einfalt~\textit{et al.}  \pub{WACV2023} \cite{DBLP:conf/wacv/EinfaltLL23}& 351 & \checkmark & 10.4 & 0.5 & 1 & 44.2/35.7 & - \\
      STCFormer  \pub{CVPR2023}  \cite{DBLP:conf/cvpr/TangQHHY23}& 243 & $\times$ & 4.7 & 19.6 & 80 & 41.0/32.0 & 21.3 \\
      STCFormer-L  \pub{CVPR2023}  \cite{DBLP:conf/cvpr/TangQHHY23}& 243 & $\times$ & 18.9 & 78.2 & 321 & 40.5/31.8 & - \\
      PoseFormerV2  \pub{CVPR23} \cite{DBLP:conf/cvpr/ZhaoZLW023} & 243 & \checkmark & 14.4 & 4.8 & 20 & 45.2/35.6 & -\\
      GLA-GCN  \pub{ICCV2023} \cite{DBLP:conf/iccv/YuZ0ZLC23}& 243 & \checkmark & 1.3 & 1.5 & 6 & 44.4/34.8 & 21.0\\
      MotionBERT   \pub{ICCV2023}  \cite{DBLP:conf/iccv/ZhuMLLW023} & 243 & $\times$ & 42.3 & 174.8 & 719 & 39.2/32.9 & 17.8\\
      HDFormer  \pub{IJCAI2023} \cite{chen2023hdformer}& 96 & $\times$ & 3.7 & 0.6 & 6 & 42.6/33.1 & 21.6\\
      MotionAGFormer-L  \pub{WACV2024} \cite{DBLP:conf/wacv/MehrabanAT24}& 243 & $\times$ & 19.0 & 78.3 & 322 & 38.4/32.5 & 17.3 \\
      KTPFormer  \pub{CVPR2024} \cite{DBLP:conf/cvpr/PengZ024} & 243 & $\times$ & 35.2 & 76.1 & 313 & 40.1/31.9 & 19.0 \\ 
      PoseMagic \pub{AAAI2025} \cite{DBLP:journals/corr/abs-2408-02922}& 243 & $\times$ & 14.4 & 20.29 & 84 & 37.5/- & - \\
       PoseMamba-S \pub{AAAI2025} \cite{huang2024posemamba}& 243 & $\times$ & 0.9 & 3.6 & 15 & 41.8/35.0 & 20.0 \\
      PoseMamba-B \pub{AAAI2025} \cite{huang2024posemamba}& 243 & $\times$ & 3.4 &13.9 & 57 & 40.8/34.3 & 16.8 \\
     PoseMamba-X \pub{AAAI2025} \cite{huang2024posemamba}& 243 & $\times$ & 26.5 & 109.9 & 452 & 37.1/31.5 & 14.8 \\
     \midrule 
      SAMA-S (Ours) & 243 & $\times$ & 1.1 & 3.9 & 16& 40.6/34.0&20.2\\
      SAMA-B (Ours) & 243 & $\times$ & 3.3 & 11.7 & 48& 37.7/32.0 & 13.6 \\
      SAMA-L (Ours) & 243 & $\times$ & 17.3 & 53.2&219 & \underline{36.9}/\underline{31.3} & \underline{11.9} \\
      SAMA-S (Ours) & 351 & $\times$ & 1.1 & 6.3 & 18 & 40.2/33.8&19.5\\
      SAMA-B (Ours) & 351 & $\times$ & 3.3 & 18.9 & 54 & 37.4/31.7 & 12.4 \\
      SAMA-L (Ours) & 351 & $\times$ & 17.3 & 82.1 & 234 & \textbf{36.5}/\textbf{31.0} & \textbf{11.4} \\
      \fontsize{7.5pt}{1em}\selectfont{\color{updatagrey}{\textit{~{vs. prev. SoTA}}}} & - & - &  \fontsize{7.5pt}{1em}\selectfont{\color{updatagrey}{\textsl{$\downarrow$\textbf{11.2}}}} & \fontsize{7.5pt}{1em}\selectfont{\color{updatagrey}{\textsl{$\downarrow$\textbf{27.8}}}} & \fontsize{7.5pt}{1em}\selectfont{\color{updatagrey}{\textsl{$\downarrow$\textbf{218}}}} & \fontsize{7.5pt}{1em}\selectfont{\color{updatagrey}{\textsl{$\downarrow$\textbf{0.6}}}}/\fontsize{7.5pt}{1em}\selectfont{\color{updatagrey}{\textsl{$\downarrow$\textbf{0.5}}}} & \fontsize{7.5pt}{1em}\selectfont{\color{updatagrey}{\textsl{$\downarrow$\textbf{3.4}}}} \\
      \bottomrule
    \end{tabular}
\vspace{-5pt}
\end{table*}

\subsection{Experimental Setup}
\label{para:Experiments - Experimental Setup}

\noindent\textbf{Datasets.} 
We conduct experiments on two widely used datasets, Human3.6M~\cite{ionescu2013human3} and MPI-INF-3DHP~\cite{mehta2017monocular}.
\begin{itemize}[leftmargin=*, itemsep=0pt, parsep=-2pt, listparindent=-10pt]
    \item \textbf{Human3.6M} is the most commonly used indoor dataset for monocular 3D human pose estimation task, containing 3.6 million human poses and corresponding images. It includes 11 subjects performing 15 daily activities. Following established protocols in recent studies~\cite{huang2024posemamba,DBLP:conf/iccv/ZhuMLLW023,DBLP:conf/iccv/ZhuMLLW023}, we take data from subjects 1, 5, 6, 7, 8 for training, and subjects 9, 11 for testing. 
    We take Mean Per-Joint Position Error  (\textcolor{citegrey}{MPJPE, $mm, \downarrow$}) and Pose-aligned MPJPE (\textcolor{citegrey}{P-MPJPE, $\%, \downarrow$}) as the main evaluation matrices. More details are in the supplementary materials.
    \vspace{2mm}
    \item \textbf{MPI-INF-3DHP} is another challenging large-scale dataset captured in both indoor and outdoor environments, comprising over 1.3 million frames from 8 subjects performing 8 activities.
    We take Mean Per-Joint Position Error (\textcolor{citegrey}{MPJPE, $mm, \downarrow$}), Percentage of Correct Keypoints (\textcolor{citegrey}{PCK, $\%, \uparrow$}) and Area Under Curve  (\textcolor{citegrey}{AUC, $\%, \uparrow$}) as the main evaluation matrices.
\end{itemize}

\noindent\textbf{Implementation details.}
Our model, is trained end-to-end, following distinct protocols 
for dataset 
as detailed below:

\begin{itemize}[leftmargin=*, itemsep=0pt, parsep=-2pt, listparindent=-10pt]
    \item \textbf{Human3.6M}: We train the model for 80 epochs using the AdamW optimizer with a batch size of 8. We set the sequence length to 351 and 243. The initial learning rate is established at 5e-5 with an exponential learning rate decay schedule, utilizing a decay factor of 0.99.
    Following previous method  \cite{huang2024posemamba,DBLP:conf/iccv/ZhuMLLW023,DBLP:journals/corr/abs-2408-02922}, we utilize SHNet \cite{xu2021graph} to extra 2D human poses and ground true input from Human3.6M for fair comparison.
    \vspace{2mm}
    \item \textbf{MPI-INF-3DHP}: Our model is trained for 90 epochs using the AdamW optimizer and the batch size is set as 16. Following the previous work \cite{{huang2024posemamba,DBLP:journals/corr/abs-2408-02922}}, the sequence length is set as 81. The initial learning rate is established at 5e-4 with an exponential learning rate decay schedule, utilizing a decay factor of 0.99. We employ the 2D ground true pose from MPI-INF-3DHP as input. 
\end{itemize}

\noindent\textbf{Baselines.} We compare our method with the state-of-the-art PoseMamba and PoseMagic.

\begin{itemize}[leftmargin=*, itemsep=0pt, parsep=-2pt, listparindent=-10pt]
   \item \textbf{PoseMamba}. 
   Utilizing a global-local spatial-temporal SSM block, PoseMamba effectively models human joint correlations, while neglecting the inherent topology and ignores motion differences among joints.

   \vspace{2mm}
   
   \item \textbf{PoseMagic}.
    Leveraging a hybrid Mamba-GCN architecture that explicitly captures the relationships between neighboring joints, PoseMagic incorporates a local enhancement module for structure modeling. Although effective at learning the underlying 3D structure, the approach uniformly treats all joints, thereby overlooking the distinct modeling requirements of joint motion.

\end{itemize}

\subsection{Indoor Monocular 3D Human Pose Estimation}
\label{para:Experiments - Indoor monocular 3D human pose estimation}
\noindent\textbf{Quantitative comparison.}
The comparative performance of various methodologies in terms of indoor monocular 3D human pose estimation is systematically listed in Tab.~\ref{tab:human3.6m-comparison}. The results unequivocally demonstrate that our proposed method exhibits superior performance, registering an exemplary state-of-the-art MPJPE score of $36.5$.

In direct comparison with the sota method PoseMamba-X~\cite{huang2024posemamba} with SAMA-L
, our method exhibits a marked enhancement of $0.6mm$ MPJPE  reduction. 
Moreover, our method consistently attains high accuracy results across settings of different sizes: $40.2mm$, $37.4mm$,  for different variants {SAMA-S} / {SAMA-B}, respectively. 
Specifically, these variants surpass PoseMamba among models with comparable parameter scales.
Furthermore, when aligning the estimated poses, our SAMA-L achieves a P-MPJPE of $31.0$, reaching the advanced level. Across different model scales, our approach consistently outperforms PoseMamba. Lastly, with ground truth 2D poses as input, our SAMA-L achieves an MPJPE of $11.4mm$, marking a significant improvement over PoseMamba ($11.4~ \textit{v.s.}~14.8$).
We attribute this to the core module of our algorithm, structure-aware state integrator and motion-adaptive state modulator. They aggregate pose topology information and adaptively model the varying motion characteristics of different joints in state space.

\noindent\textbf{Efficiency comparison.}
To showcase the efficiency of our method, we compare it with others in terms of parameter count and MACs per frame. Especially, our SAMA-B uses only 3.3M parameters (1/2 of PoseMamba-L) and 54M MACs per frame (less than half of PoseMamba-L). On the dataset with SHNet-detected 2D poses, it achieves $0.7mm$ lower prediction error than PoseMamba-L. When using 2D ground truth as input, it surpasses all previous models. 
Additionally, our SAMA-L achieves significantly lower parameter count and MACs per frame compared to the previous SOTA PoseMamba-X while maintaining superior accuracy, reducing the prediction error by 0.6mm with SHNet-detected 2D poses and 3.4mm with ground truth input.
We attribute this to our module's structure-aware joint feature fusion and state fusion, which are based on a lightweight learnable adjacency matrix. Additionally, in MSM, we leverage basic functions to identify joint motion characteristics without introducing excessive computation.

\subsection{In-the-wild 3D Human Pose Estimation}
\label{para:Experiments - In-the-wild pose estimation}
To evaluate  robustness, we compare our SAMA's performance with other methods in \cref{tab:mpi} on MPI-INF-3DHP, which contains in-the-wild scenario. 
For a fair comparison,
we follow the previous works \cite{DBLP:journals/corr/abs-2408-03540,DBLP:journals/corr/abs-2408-02922,DBLP:conf/wacv/MehrabanAT24} to take the ground true 2D keypoints as
input and the sequence length is set as 81.
Our SAMA achieves state-of-the-art performance with an MPJPE of 14.4 mm, compared to the previous best method, PoseMamba. Additionally, our method surpasses PoseMagic in terms of AUC and PCK by 0.2\% and 0.7\%, respectively. 
These results demonstrate the robustness of our method on the outdoor dataset MPI-INF-3DHP, while maintaining strong performance even with short sequences.

\begin{table}[!t]
    \centering
    \setlength{\tabcolsep}{3.0pt}
    \renewcommand\arraystretch{1.0}
    \setlength{\abovecaptionskip}{5pt}  
    \setlength{\belowcaptionskip}{5pt}
    \caption{Quantitative comparisons on MPI-INF-3DHP dataset. The best performances are \textbf{bold}.
    MPJPE($mm, \downarrow$), PCK($\%, \uparrow$) and AUC($\%, \uparrow$) are reported.
    T denotes the number of input frames.}
    \label{tab:mpi}
    \small

    \begin{tabular}{lcccc}
    \toprule
     Method& T & PCK$\uparrow$ &AUC $\uparrow$ &MPJPE $\downarrow$\\
    \midrule
    Anatomy3D~\pub{TCSVT2021}~\cite{chen2021anatomy} &81 &87.8&53.8&79.1\\
    PoseFormer~\pub{ICCV2021}~\cite{zheng20213d}& 9  &88.6&56.4&77.1 \\
    MixSTE~\pub{CVPR2022}~\cite{DBLP:conf/cvpr/Zhang0YCY22}&27&94.4&66.5&54.9 \\
    MHFormer~\pub{CVPR2022}~\cite{DBLP:conf/cvpr/Li00WG22}&9 &93.8&63.3&58.0\\
    P-STMO~\pub{ECCV2022}~\cite{DBLP:conf/eccv/ShanLZWMG22}&81   &97.9&75.8&32.2\\
    GLA-GCN~\pub{ICCV2023}~\cite{DBLP:conf/iccv/YuZ0ZLC23}&81 &98.5&79.1&27.8 \\
    STCFormer~\pub{CVPR2023}~\cite{DBLP:conf/cvpr/TangQHHY23}&81 &98.7&83.9&23.1 \\
    PoseFormerV2~\pub{CVPR2023}~\cite{DBLP:conf/cvpr/ZhaoZLW023}&81  &97.9&78.8&27.8 \\
    MotionAGFormer~\pub{WACV2024}~\cite{DBLP:conf/wacv/MehrabanAT24} &81&98.2&85.3&16.2\\
    KTPFormer~\pub{CVPR2024}~\cite{DBLP:conf/cvpr/PengZ024}&81 &98.9&85.9&16.7 \\
    PoseMagic~\pub{AAAI2025}~\cite{DBLP:journals/corr/abs-2408-02922} &81&98.8&87.6&14.7\\
    PoseMamba~\pub{AAAI2025}~\cite{DBLP:journals/corr/abs-2408-03540}&81&-&-&14.5\\
    \midrule
    SAMA (Ours)&81           &\textbf{99.0}&\textbf{88.3}&\textbf{14.4} \\
    \bottomrule
\end{tabular}
\vspace{-10pt}
\end{table}

\begin{figure*}[t!]
\begin{minipage}[t!]{0.36\linewidth}
\centering
\small
\setlength{\tabcolsep}{8.5pt}
\renewcommand\arraystretch{1.0}
\setlength{\abovecaptionskip}{0pt}  
\setlength{\belowcaptionskip}{0pt}
\tabcaption{Ablation of the main components in our method.}
\label{tab: ablation: main components}
    \begin{tabular}{cccc}
    \toprule
    Vanilla SSD & SSI & MSM & MPJPE \\
    \hline
     - & - & -&39.9 \\
     \checkmark & - & -&39.3 \\
    \checkmark & \checkmark & -&38.4 \\
    \checkmark & - &\checkmark &38.5 \\
    \checkmark & \checkmark & \checkmark &\textbf{37.4} \\
    \bottomrule
    \end{tabular}
\end{minipage}
\qquad
\hspace{-0.7cm}
\begin{minipage}[t!]{0.32\linewidth}
\centering
\small
\setlength{\tabcolsep}{8.5pt}
\renewcommand\arraystretch{1.5}
\setlength{\abovecaptionskip}{0pt}  
\setlength{\belowcaptionskip}{16pt}
\tabcaption{Comparison with other spatial scanning methods.}  
\label{tab: abaltion: scaning}
    \begin{tabular}{l c c}
    \toprule
    Spatial Learning  & MPJPE &MACs \\
    \hline
    bi-direction~\cite{DBLP:journals/corr/abs-2408-02922}   & 38.2 & 58.12 \\
    global-local~\cite{DBLP:journals/corr/abs-2408-03540} & 37.9  & 58.12 \\
    vanilla + SSI (Ours) & \textbf{37.4} & 53.95 \\
    \bottomrule
    ~ & ~ & ~ \\
    \end{tabular}
\end{minipage}
\qquad
\hspace{-0.2cm}
\begin{minipage}[t!]{0.25\linewidth}
\centering
\small
\setlength{\tabcolsep}{10pt}
\renewcommand\arraystretch{1.5}
\setlength{\abovecaptionskip}{0pt}  
\setlength{\belowcaptionskip}{16pt}
\tabcaption{Effect of different motion detection function.} 
\label{Tab:motioncap}
    \begin{tabular}{l c}
    \toprule
    Motion Learning  & MPJPE \\
    \hline
    Baseline    & 38.4  \\
    Linear & 38.0 \\
    Point-wise Conv & \textbf{37.4}   \\
    \bottomrule
    ~ & ~ \\
    \end{tabular}
\end{minipage}
\vspace{-1.5cm}
\end{figure*}

\begin{figure*}[t!]
\begin{minipage}[t!]{0.29\linewidth}
\centering
\small
\setlength{\tabcolsep}{0pt}
\renewcommand\arraystretch{1.1}
\setlength{\abovecaptionskip}{0pt}  
\setlength{\belowcaptionskip}{5pt}
\vspace{-0.1cm}
\tabcaption{Generalization of our algorithm.}
\label{Tab: ablation: generalization}
    \begin{tabular}{l c}
    \toprule
    Method  & MPJPE   \\
    \midrule
    MixSTE\pub{CVPR2022}~\cite{DBLP:conf/cvpr/Zhang0YCY22} & 40.9   \\
    MixSTE + $\textbf{\texttt{Ours}}$ & 40.3 \fontsize{7.5pt}{1em}\selectfont{\color{updatagreen}{\textsl{$\downarrow$\textbf{0.6}}}} \\
    \hdashline
    MotionBERT\pub{ICCV2023}~\cite{DBLP:conf/iccv/ZhuMLLW023}& 39.2 \\
    MotionBERT + $\textbf{\texttt{Ours}}$ & 38.0 \fontsize{7.5pt}{1em}\selectfont{\color{updatagreen}{\textsl{$\downarrow$\textbf{1.2}}}}   \\
    \hdashline
    MotionAGFormer\pub{WACV2024}~\cite{DBLP:conf/wacv/MehrabanAT24} &38.4 \\
    MotionAGFormer + $\textbf{\texttt{Ours}}$ & 37.5 \fontsize{7.5pt}{1em}\selectfont{\color{updatagreen}{\textsl{$\downarrow$\textbf{0.9}}}} \\
    
    \bottomrule
    \end{tabular}
\end{minipage}
\qquad
\hspace{-0.8cm}
\begin{minipage}[t!]{0.34\linewidth}
\setlength{\abovecaptionskip}{5pt}  
\setlength{\belowcaptionskip}{5pt}
\vspace{+0.3cm}
\figcaption{Visual comparable results of estimated 3D poses between PoseMamba and ours.}
  \label{fig:vis_comparison}
  \includegraphics[width=0.99\linewidth]{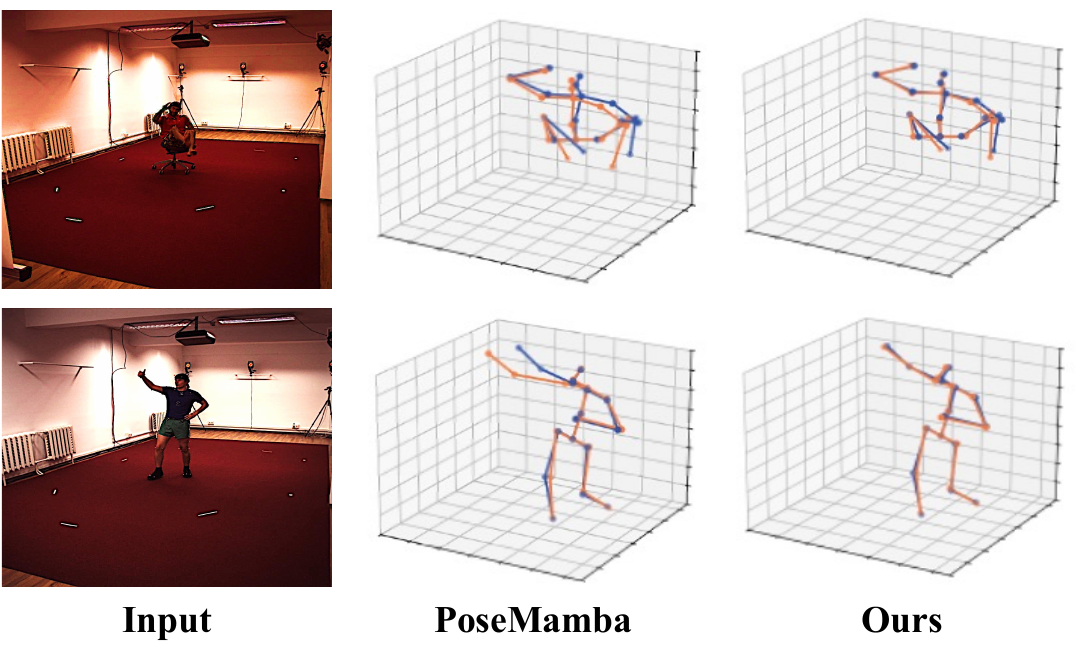}   
\end{minipage}
\qquad
\hspace{-0.7cm}
\begin{minipage}[t!]{0.36\linewidth}

\setlength{\abovecaptionskip}{0pt}  
\setlength{\belowcaptionskip}{20pt}
\figcaption{Statistical motion intensity and timescale $\Delta$ results across different joints.}

  \includegraphics[width=1.0\linewidth]{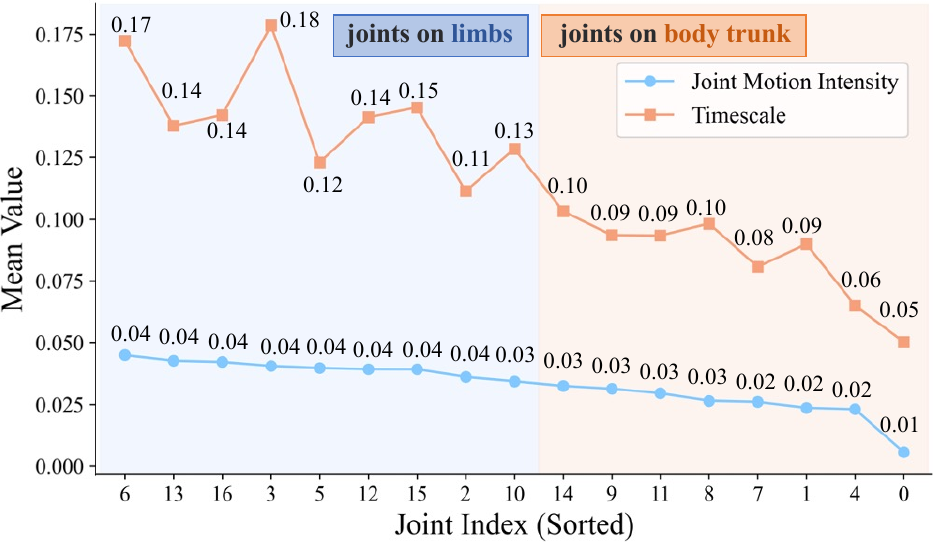}   
  \label{visdt}
\end{minipage}
\vspace{-1.0cm}
\end{figure*}

\subsection{Ablation Study}
\label{para:Experiments - Ablation Study}
We conduct a series of ablation study on Human3.6M~\cite{ionescu2013human3} to validate the efficacy of our core algorithm designs with our SAMA-B as the base model.

\noindent\textbf{Effect of our main components.}
To evaluate the impact of our core algorithm, we conducted an analysis by removing the structure-aware state integrator and motion-adaptive state modulator.
As presented in \cref{tab: ablation: main components}, the baseline model, composed of stacked blocks without our proposed components, achieves an MPJPE of $39.9$ mm.
Incorporating the basic SSD module leads to a $0.6$ mm reduction in MPJPE.
Built on this setting, SSI yields a performance improvement to $38.4$ mm, attributed to its ability to enhance joint correlation modeling via a learnable adjacency matrix in the state space.
Besides, MSM further improves performance to $37.4$ mm, owing to its capability to adaptively capture motion patterns by controlling the timescale.
The results also verify that combining SSI and MSM yields the best results, indicating the effectiveness of considering topology information aggregation in space and different joint motion characteristics in time.

\noindent\textbf{Generalization evaluation.}
To evaluate the generalization capability of our approach, we integrate our core discrete joint modeling component into other methods.
Specifically, we prepend our SSI and MSM to the networks without modifying the remaining architecture.
For a fair comparison, we adopt their default implementation settings, including hyperparameters and augmentation strategies.
\cref{Tab: ablation: generalization} presents the comparative results on Human3.6M.
As observed, our approach significantly enhances the performance of the baseline estimation networks, achieving reductions of $0.6$, $1.2$, and $0.9$ mm in MPJPE for MixSTE~\cite{DBLP:conf/cvpr/Zhang0YCY22}, MotionBERT~\cite{DBLP:conf/iccv/ZhuMLLW023}, and MotionAGFormer~\cite{DBLP:conf/wacv/MehrabanAT24}, respectively.
These consistent performance improvements illustrate the wide potential benefit of our algorithm. Notably, `MotionAGFormer$_{\!}$ +$_{\!}$ \texttt{Ours}' achieves an MPJPE of $37.5$ mm, which is on par with the advanced methods.  
This result is particularly impressive, considering the fact that the improvement is solely achieved by integrating our module, without any additional modifications.
The success of our approach can be attributed to the fact that our algorithm not only effectively complements the topological connections between joints but also takes into account the distinct motion characteristics of different joints, further enhancing overall performance.

\noindent\textbf{Comparison with various spatial learning methods.} 
To demonstrate the effectiveness of our SSI, we replaced the spatial dependency learning part of our model with the previous methods, bi-directional scanning method in PoseMagic and global-local scanning method in PoseMamba.
The bi-directional scanning method sequentially processes joint indices in both descending and ascending orders, thereby neglecting the intrinsic connectivity among joints.  
Besides, the global-local strategy employs a predefined local motion-specific scanning pattern, which yields only marginal performance gains at the expense of considerable computational cost.  
As shown in \cref{tab: abaltion: scaning}, our approach, which integrates a simple vanilla scanning method with the SSI, achieves the best MPJPE result of $37.4$ mm with lower computational cost, demonstrating greater efficiency compared to the more complex scanning strategies.
This result underscores the capability of our SSI in effectively capturing dynamic spatial joint dependencies.

\noindent\textbf{Effect of motion-adaptive state modulator.} 
We visualize the effect of motion-adaptive state modulator in \cref{visdt}. MSM leverages the motion characteristics between adjacent frames to learn a timescale that dynamically balances the influence of the previous state and the current input for the current frame's output, thereby capturing richer joint motion features.
As shown in the figure, joints on limbs (\textit{e.g.}, joint 3, 6, 13, 16, 5 and 12), which exhibit greater average motion intensity, correspond to larger timescales, while joints on the body trunk (\textit{e.g.}, joint 0, 1, 4, 7 and 8), which move less, correspond to smaller timescales.
This correlation between motion intensity and timescale confirms the fundamental rationale behind our design of motion-adaptive state modulator. Specifically, our model leverages motion information so that larger motion amplitudes correspond to larger timescales. This allows the model to reduce reliance on the previous state when encountering intense motion, preventing it from erroneously smoothing the motion trajectory in such cases.

\noindent\textbf{Effect of motion capture method.} 
We explore two simple functions to capture motion cues between adjacent joints to regulate the timescale, using SAMA-B without motion capturing as the baseline, as shown in \cref{Tab:motioncap}.
Point-wise convolution (1D conv, kernel size 2) captures local motion patterns, enabling dynamic timescale adjustments. A simple linear layer preserves complete adjacent joint features, enhancing joint dependency modeling.
Both methods use zero padding on the left and improve performance, demonstrating the effectiveness of joint-specific motion information in regulating timescales.
In practical applications, we adopt point-wise convolution for implementation.

\noindent\textbf{Visualization of estimated poses.}
\cref{fig:vis_comparison}  illustrates the 3D pose predictions of PoseMamba and our method, where \textcolor{gtblue}{blue} / \textcolor{orange}{orange} denotes the ground \textcolor{gtblue}{truth} / \textcolor{orange}{estimated poses}, respectively. It reveals that the estimated poses generated by our approach demonstrate superior accuracy compared to those of PoseMamba, particularly in the highly dynamic limb regions. This highlights the effectiveness of our joint-specific modeling strategy, enabling more precise motion capture and consequently enhancing overall performance.

\section{Conclusion}
In this work, we introduce a new algorithm tailored for lifting-based pose estimation.
Our algorithm incorporates a structure-aware and motion-adaptive strategy, facilitating dynamic joint connection modeling and personalized motion adaptation, enabling more precise motion trajectory reconstruction while preserving intrinsic motion characteristics, thereby ensuring enhanced representation of joint dependencies. 
Experimental evaluations on comprehensive benchmarks manifest its superiority in accuracy and efficiency with reduced computational cost.

{
    \small
    \bibliographystyle{ieeenat_fullname}
    \bibliography{main}
}

\end{document}